\pdfoutput=1
\documentclass[11pt]{article}

\usepackage[]{naacl2021}

\usepackage{times}
\usepackage{latexsym}

\usepackage{graphicx}
\usepackage{subcaption}
\usepackage{multirow}
\usepackage{amsmath}
\usepackage{diagbox}

\usepackage{url}
\usepackage{xcolor}
\usepackage{todonotes} % insert [disable] to disable all notes

\DeclareMathOperator*{\argmax}{arg\,max}
\newcommand\numberthis{\addtocounter{equation}{1}\tag{\theequation}}

\usepackage[T1]{fontenc}
\usepackage[utf8]{inputenc}

\usepackage{microtype}

\title{Learning Feature Weights using Reward Modeling for Denoising Parallel Corpora}
\author{Gaurav Kumar, Philipp Koehn, Sanjeev Khudanpur \\
  Center for Language and Speech Processing \\
  Johns Hopkins University \\
  {\tt gkumar@cs.jhu.edu},
  {\tt \{phi, khudanpur\}}\tt{@jhu.edu} \\
  }

\begin{document}
\maketitle
\begin{abstract}
Large web-crawled corpora represent an excellent resource for improving the performance of Neural Machine Translation (NMT) systems across several language pairs. However, since these corpora are typically extremely noisy, their use is fairly limited. Current approaches to dealing with this problem mainly focus on filtering using heuristics or single features such as language model scores or bi-lingual similarity. This work presents an alternative approach which learns weights for multiple sentence-level features.
These feature weights which are optimized directly for the task of improving translation performance,
are used to score and filter sentences in the noisy corpora more effectively. We provide results of applying this technique to building NMT systems using the Paracrawl corpus for Estonian-English and show that it beats strong single feature baselines and hand designed combinations. 
Additionally, we analyze the sensitivity of this method to different types of noise and  explore if the learned weights generalize to other language pairs using the Maltese-English Paracrawl corpus. 
\end{abstract}

\section{Introduction}
Large parallel corpora such as Paracrawl \cite{banon-etal-2020-paracrawl} which have been crawled from online resources hold the potential to drastically improve performance of neural machine translation systems across both low and high resource language pairs. However, since these extraction efforts mostly rely on automatic language identification and document/sentence alignment methods, the resulting corpora are extremely noisy. The most frequent noise types encountered are sentence alignment errors, wrong language in source or target, and untranslated sentences. As outlined by \citet{khayrallah-koehn-2018-impact}, training algorithms for neural machine translation systems are particularly vulnerable to these noise types. As such, these web-crawled corpora have seen limited use in training large NMT systems. 

This paper proposes a method for denoising and filtering noisy corpora which explores and searches over weighted combinations of features. During NMT training, we score sentences and create batches using random weight vectors. These batches are use to train the system and measure improvement over the validation set (reward). Finally, by modeling the weight-\emph{reward} function, we learn the set of weights which maximize reward and are used to score and filter the noisy dataset. At a high level, this method
    (i)~allows the use of multiple sentence level features,
    (ii)~learns a set of interpolation weights for the features which directly maximize translation performance,
    (iii)~requires no prior knowledge about which features are informative or even if they are mutually redundant, and
    (iv)~trains within the NMT pipeline and does not require any special infrastructure.

We include experiments which apply this method to building NMT systems for the noisy Estonian-English Paracrawl dataset and show that it beats strong single feature filtering-baselines and hand-designed feature interpolation. Additionally, we analyze the robustness of this method in the presence of specific kinds of noise \cite{khayrallah-koehn-2018-impact} via a controlled experiment on the Europarl datasets. Finally, we look at the impact of transferring the learned weights from one language pair (Estonian-English) to a noisy dataset of another language pair (Maltese-English Paracrawl). 

We present related work in Section~\ref{sec:related-work}. Section~\ref{sec:methods} describes the procedure we use to model and search over the weight-feature-\emph{reward} space to estimate feature weights which maximize translation performance. Our experiment design, datasets and features, appear in Section~\ref{sec:experiment}. Section~\ref{sec:results} includes our primary results where we compare the performance of the proposed method to strong single feature filtering baselines and hand-design feature weights. We conclude in section~\ref{sec:analysis} with an analysis of this method's performance at filtering specific kinds of noise and the application of learned weights to a different language pair.

\section{Related Work} \label{sec:related-work}
Existing efforts towards filtering and denoising noisy corpora focus on pre-filtering using hand-crafted rules and by using sentence pair scoring and filtering methods. Deterministic hand-crafted rules \cite{hangya-fraser-2018-unsupervised, kurfali-ostling-2019-noisy} remove sentence pairs with extreme lengths, unusual sentence length ratios and exact source-target copies, and are extremely effective in removing most of the obvious automatic extraction errors. Automatic sentence pair scoring functions have been used successfully to filter noisy corpora as well. This includes the use of language models \cite{rossenbach-etal-2018-rwth}, neural language models trained on trusted data \cite{junczysdowmunt2019dual} and lexical translation scores \cite{gonzalez-rubio-2019-webinterpret}. \citet{chaudhary-etal-2019-low} propose the use of cross-lingual sentence embeddings for determining sentence pair quality while several efforts \cite{kurfali-ostling-2019-noisy, soares-2019-unsupervised, bernier-colborne-lo-2019-nrc} have focused on the use of monolingual word embeddings. \citet{parcheta-etal-2019-filtering} use a machine translation system trained on clean data to translate the source sentences of the noisy corpus and evaluate the translation against the original target sentences using BLEU scores. \citet{erdmann-gwinnup-2019-quality} and \citet{sen-etal-2019-parallel} propose similar methods using METEOR scores and Levenshtein distance respectively. \citet{rarrick2011mt}, \citet{venugopal-etal-2011-watermarking} and \citet{antonova-misyurev-2011-building} present techniques for detecting machine translated sentence pairs in corpora. Tools such as LASER \cite{schwenk-douze-2017-learning}, BiCleaner \cite{prompsit:2018:WMT} and Zipporah \cite{xu-koehn-2017-zipporah} have been used \cite{chaudhary-etal-2019-low} for noisy corpus filtering. Curriculum learning has been used  to obtain policies for data selection which can expose the model to noisy samples less often during training \cite{wang-etal-2018-denoising, kumar-etal-2019-reinforcement}. More recently, \citet{elnokrashy2020score} and \citet{esplgomis20a}
% \footnote{\url{www.statmt.org/wmt20/program.html}}
have used classifier based approaches to filtering noisy parallel data. 
\section{Method} \label{sec:methods}
\begin{figure}[t]
    \centering
    \includegraphics[width=\linewidth]{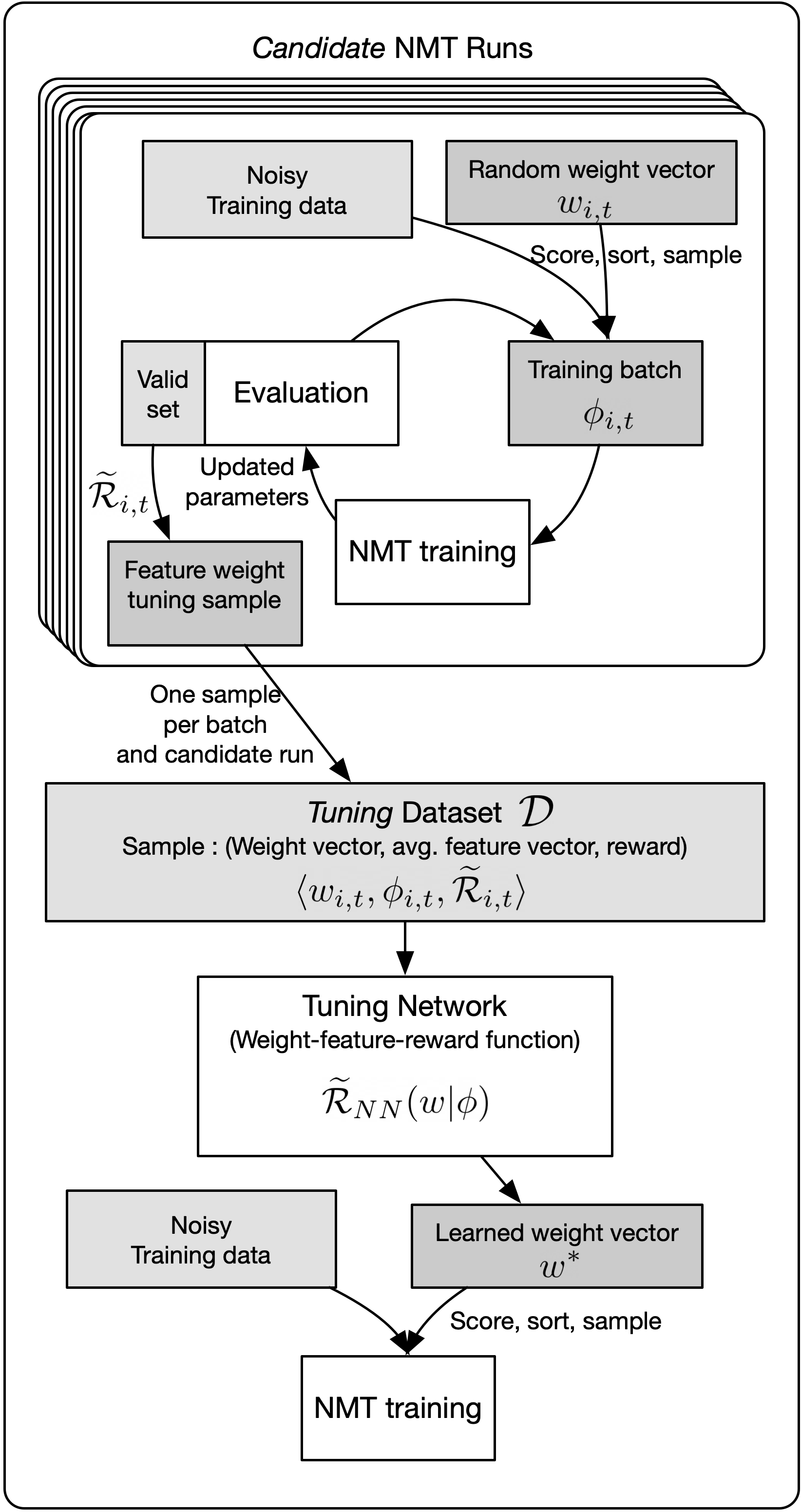}
    \caption{Overview of the proposed method for learning weights for sentence-level features to filter noisy parallel data and improve translation performance.}
    \label{fig:overview}
\end{figure}

The proposed method centres around finding weights for combining sentence-level features, which are then used to compute sentence-level scores and filter the noisy corpus. While the choice of features can be arbitrary, this method's performance will eventually depend on their quality, and we would ideally want them to be informative and decorrelated.

Figure~\ref{fig:overview} provides an overview of the proposed method.  We first train a number of \emph{candidate}
neural machine translation (NMT) systems.  During training for each candidate system, we repeatedly (i)~generate a random weight vector, (ii)~sample a batch of sentences from the noisy corpus based on sentence-level scores computed using this weight vector, (iii)~update NMT system parameters using this batch, and (iv)~measure the improvement in translation quality on a validation set following this update.  
The weight vector $w$, the average feature vector $\phi$ of the batch, and the improvement ${\cal R}$ on the validation set (\emph{reward}) are recorded for each batch $t$ during the training of each candidate NMT system $i$, and $\langle w_{i,t},\phi_{i,t},{\cal R}_{i,t}\rangle$ becomes a sample in new data set ${\cal D}$,
called the \emph{tuning} data set\footnote{Not to be confused with the validation set which contains sentence pairs, this dataset is solely used to model the weight-reward function and contains no sentence identity beyond feature vectors.}, for learning feature weights to maximize reward. Hence, even though the parameters of the \emph{candidate} systems are not used directly, they are used to gather noisy \emph{candidate} evaluations of the latent weight-feature-reward function.

Once we have ${\cal D}$, we use a feed-forward network to learn the weight vector that maximizes the reward. The learned weight vector $w^\ast$ is then used to compute sentence-level scores and filter the noisy data set. The \emph{final} NMT system is trained using this \emph{clean} data set.

Some subtleties in normalizing the observed rewards and learning weights are explained below.

\subsection{Candidate NMT runs} \label{sec:candidate}

Note from the bottom of Figure \ref{fig:overview} that the learned weight vector $w^\ast$ is used to sort all the sentences in the noisy training data, and the top-scoring ones are used for final NMT training.
The purpose of the candidate NMT training runs is to generate the \emph{tuning} data set ${\cal D}$ from which $w^\ast$ is learned.  Therefore, the setup for the candidate runs mimics typical NMT training, but for the following differences.
\begin{enumerate}
    \item \textbf{Selecting batches}: For selecting sentences to constitute a batch, we first sample a random weight vector $w$ of dimension $|\phi|$, the number of sentence-level features, uniformly\footnote{The range of the uniform distribution represents the plausible range of weights given the features.} from $[-2.5, 2.5]^{|\phi|}$. Ideally, we would score \emph{all} sentences in the noisy data set and then filter the top sentences to create a batch.  However, this is prohibitively slow to do for every batch. Hence, we randomly sample twice the number of sentences required to constitute the batch, score them, and select the top half. For the $i^\text{th}$ sentence, the score $s_i$ is a dot product of its feature vectors with the weight vector:
    \begin{align}
        s_i = \sum_{i=1}^{|\phi|} w_i \phi_i
        \label{eqn:score-dot}
    \end{align}
    
    The selected sentences are removed from the training pool for this epoch.
    This method of batch selection ensures that the sampled weight vector determines which sentences are selected and that their average feature vector is significantly different from one obtained using unbiased/random selection. 
    
    \item \textbf{Reward computation}: The reward must represent how the choice of $w$ (through the sentences selected to form the batch) impacts translation performance. This is approximated by computing the perplexity of a validation set following a parameter update with the selected batch. However, since perplexity naturally decays in standard NMT training, batches at the beginning of the training will naturally receive larger rewards, obscuring the impact of sentence selection.  We mitigate this effect by using delta-perplexity, i.e. the change in perplexity of the validation set over a window of updates. 
    \item \textbf{Accumulating training samples}: For each batch $t$ of candidate run $i$, we collect the random weight vector $w_{i,t}$,
    the batch feature vector $\phi_{i.t}$, defined as the average of the feature vectors of all sentences in the batch, and the reward ${\cal R}_{i,t}$. These triples are gathered from all batches during training, across all candidate training runs, to form the data set ${\cal D}$ for learning the feature weights.
\end{enumerate}

\subsection{Reward Normalization}
As a further way to make the rewards time-invariant with respect to NMT training, the observed rewards ${\cal R}_{i.t}$ are normalized with respect to an expected reward estimated from a set of \emph{baseline} NMT runs.  Specifically, at each time step $t$, we compute the rewards $\mathcal{R}^b_{j,t}$ of $j=1,\ldots,J$ concurrent training runs---whose batches selected in the standard manner---and, for each of the candidate NMT runs, we set
\begin{align}
\widetilde{\mathcal{R}}_{i,t} = \mathcal{R}_{i,t} - \frac{1}{J} \sum_{j=1}^J \mathcal{R}^b_{j,t},
\end{align}
where $J$ is the number of baseline systems used.

Going forward, we do not need to track the identity of the update which led to a training sample, $t$, or the candidate system $c_i$ which produced it.

\subsection{Learning Feature Weights}
The $i^\text{th}$ sample $\langle w_{i},\phi_{i},\widetilde{\cal R}_{i}\rangle$ in $\mathcal{D}$ may be viewed as a (noisy) evaluation of an unknown function $\mathcal{R}(w|\phi)$. This function maps a vector $w$ to final NMT quality, given a fixed sentence-level feature function $\phi$ and the stipulation that sentences are selected for training based on a weighted combination of their feature values using weights $w$.
Furthermore, if we learn this function using $\mathcal{D}$, we may use the $w^\ast$ that maximizes the learned function $\widetilde{\mathcal{R}}_{NN}(w | \phi)$ for our final denoising and NMT training. Specifically, we propose to use
\begin{align*}
    w^\ast =& \argmax_w \mathcal{R}(w|\phi)\\
    \approx& \argmax_w  \widetilde{\mathcal{R}}_{NN}(w | \phi) \numberthis
    \label{eqn:max-weight}
\end{align*}
We propose learning $\widetilde{\mathcal{R}}_{NN}(w|\phi)$ via a simple feed-forward neural network that maps the weights $w_{i}$ to the observed reward $\widetilde{\mathcal{R}}_{i}$.  We consider two ways of providing input to this neural network, one that uses only the $w_{i}$, and another that modulates $w_{i}$ with batch quality, represented by $\phi_{i}$. 
\begin{enumerate}
    \item \textbf{Weight-based}: We use a feed-forward network with the weight vectors $w_{i}$ as input and learn to predict the observed reward $\widetilde{\mathcal{R}}_{i}$. Since the weight vectors interact directly with the feature vectors to determine which sentences are sampled to create a batch, we hypothesize that maximizing this weight-reward function will produce feature weights which will lead to better sentence sampling.
    \item \textbf{Feature-based}: Since the \emph{tuning} samples are noisy evaluations of the function $\mathcal{R}(w|\phi)$, we often encounter samples where weight vectors are close in weight space but have different rewards. To counter this problem, when using a feed-forward network to learn $\widetilde{\mathcal{R}}_{NN}(w|\phi)$, we scale the weight vector input $w_{i}$ by the sum of the corresponding feature vector $\phi_{i}$. This has the effect of keeping weight vectors which have similar feature vectors close in input space and moving apart those with significantly different feature vectors. 
\end{enumerate}
Once this neural network is learned from $\mathcal{D}$, we perform a grid search over its input space, as defined in Section \ref{sec:candidate}, to find the maximizer of (\ref{eqn:max-weight}).

\subsection{Re-sampling and training}
The weight vector $w^\ast$ learned from the previous section is used to score all sentences from the original noisy data set.
We sort the sentences by these scores and sample the top candidates to form the \emph{clean} training data set and use it to train a standard NMT system. 

\section{Experiment Setup} \label{sec:experiment}
We use Fairseq \cite{ott2019fairseq} for our neural machine translation systems configured to be identical to the systems described in \citet{ng-etal-2019-facebook}. The feed-forward network used to tune weights has two 512-dimensional layers and is trained using standard SGD using a learning rate of 0.1. The grid search for the weights was done on the range $[-2.5, 2.5]$ with 5000 points uniformly distributed per dimension. The number of samples used for reward normalization was 3 and the window for computing the delta-perplexity reward was set to 3.

\subsection{Corpora}
We use the Paracrawl Benchmarks \cite{banon-etal-2020-paracrawl} data set in Estonian-English for all our experiments.
These consist of documents where sentences were aligned using Vecalign \cite{thompson-koehn-2019-vecalign} and then de-duplicated so that each sentence pair only occurs once in the data set. The test and validation sets for our experiments in Estonian-English are \emph{newstest2018} and \emph{newsdev2018} respectively. Statistics of these corpora appear in Table~\ref{tab:paracrawl-stats}.

% \begin{table}[t!]
\begin{table}[h]
    \centering
    \begin{tabular}{|c|c|c|c|} \hline
         & \textbf{train} & \textbf{valid} & \textbf{test} \\ \hline
        %  \multicolumn{3}{|c|}{\bf{Uniform baselines}} \\ \hline
         Sentence Pairs & 22.8m & 2k & 2k \\ \hline
         Source Tokens & 190m & 29k & 31k \\ \hline 
         Target Tokens & 207m & 38k & 40k \\ \hline 
         Avg. Len (src) & 9.8 & 14.5 & 15.3 \\ \hline
         Avg. Len (tgt) & 10.7 & 19.1 & 20.1 \\ \hline
        %  Avg. Len ratio & & 0.78 & 0.78 \\ \hline
    \end{tabular}
    \caption{Statistics for the processed Estonian-Engligh (Es-En) Paracrawl data set and its corresponding validation and test sets. The training corpus was filtered using Vecalign scores; the raw corpus contains about 168m sentence pairs.}
    \label{tab:paracrawl-stats}
\end{table}

\subsection{Features} \label{ssec:features}
We use five sentence-level features for all our filtering experiments. They are, (i) IBM Model 1 alignment scores \cite{brown-etal-1993-mathematics}, (ii and iii) source and target language model scores, (iv) dual conditional cross entropy \cite{junczysdowmunt2019dual} and (v) sentence length ratio. We experimented with aggregate features such as Zipporah \cite{xu-koehn-2017-zipporah}, BiCleaner \cite{prompsit:2018:WMT} and bilingual features such as LASER \cite{schwenk-douze-2017-learning} and these were used to replicate the baselines from \citet{banon-etal-2020-paracrawl} for our dataset. 
% These features are instead used to provide filtering baselines. 
The IBM Model 1 scores were obtained using the Moses \cite{Koehn:2007:MOS:1557769.1557821} pipeline. The Estonian and English language models were trained on their respective NewsCrawl data sets\footnote{\url{statmt.org/wmt18/translation-task.html}}. The \emph{clean} machine translation model for computing the conditional dual-cross entropy scores is trained on the \emph{Europarl}v8 data set\footnote{\url{statmt.org/europarl/}}. All features are gaussianized using the Yeo-Johnson power transformation and then normalized to have zero mean and unit variance. 

\section{Results} \label{sec:results}

For our experiments, we scored all sentences in the noisy corpus, sorted and sampled the top parallel sentences to form subsets with 10, 15 and 20  million English words.
% \philipp{Sentences or Words? I think these are word counts.}
These filtered data sets were used to train standard NMT systems and performance was evaluated on the test set described in the previous section. The results of these filtering experiments appear in Table \ref{tab:results}.

First, we evaluate the efficacy of all the features we use for our interpolation task by filtering the data set on these features alone. Additionally, to include some strong baselines, we use three out-of-the-box, scoring features which provided strong results in the WMT 2020 parallel corpus filtering task\footnote{\url{statmt.org/wmt20/parallel-corpus-filtering.html}} \cite{banon-etal-2020-paracrawl, chaudhary-etal-2019-low}. These are BiCleaner, Zipporah and LASER. Of these, LASER provides the strongest filtering and translation results beating the other two by 0.3 to 0.9 BLEU points. Of the five features we use for our experiments, dual cross-entropy \cite{junczysdowmunt2019dual} is the strongest feature and matches the performance of LASER. Using source or target language model scores in isolation leads to the weakest translation performance while IBM Model 1 scores perform only slightly better than them. Surprisingly, the simple sentence length ratio feature beats all other features except dual cross-entropy by 1.4 to 1.6 BLEU points. This is a strong indicator of the type of noise in the data set and that bilingual features (even simple ones) perform better than monolingual features such as language model scores. 

\begin{table}[ht!]
    \centering
    \begin{tabular}{|c|c|c|c|} \hline
        %  & \multicolumn{3}{c||}{\bf{Estonian-English}}  \\ \hline
         & \bf{10m} & \bf{15m} & \bf{20m} \\ \hline
         \multicolumn{4}{|c|}{\bf{1-Feature Filtering Baselines}} \\ \hline
         Zipporah & 20.4 & 21.3 & 21.3 \\ \hline
         BiCleaner & 19.8 & 20.9 & 21.2 \\ \hline 
         LASER & 21.7 & 22.4 & 22.5 \\ \hline 
         IBM Model 1 & 18.1 & 19.9 & 20.8 \\ \hline 
         Target LM & 17.6 & 19.5 & 20.4 \\ \hline
         Source LM & 17.4 & 19.4 & 20.4 \\ \hline
         Dual Cross-Entropy & 21.5 & 22.4 & 22.6 \\ \hline
         Sentence Length Ratio & 19.7 & 20.2 & 21.2 \\ \hline\hline
         \multicolumn{4}{|c|}{\bf{Filtering using Feature Weights}} \\ \hline
         Uniform weight baseline & 20.9 & 21.5 & 21.6 \\ \hline
         Weight based (14) & 22.1 & \textbf{23.1} & \textbf{23.5} \\ \hline
         Feature based (15) & \textbf{22.4} & \textbf{23.1} & \textbf{23.6} \\ \hline
    \end{tabular}
    \caption{BLEU scores for the Estonian-English NMT systems where the training data was filtered using single features or a learned weighted combination of features. Feature weights were learned using the proposed method. The number of candidate runs which produced the best results appear in parentheses.
    }
    \label{tab:results}
\end{table}

\begin{figure}[t]
    \centering
    \includegraphics[width=\linewidth]{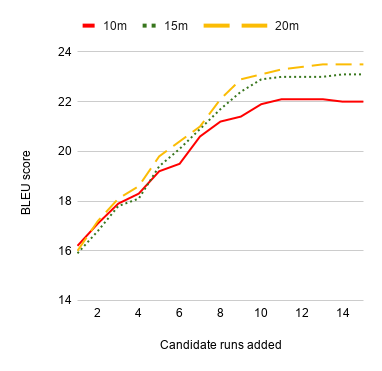}
    \caption{Improvement in BLEU scores of the final NMT system as data from additional `candidate` training runs is added to the tuning stage to learn weights. Training data was filtered using the learned weights.}
    \label{fig:candidate}
\end{figure}

Next, we look at interpolation of features using weights learned using the proposed method. As a baseline, we also include an experiment which filters based on a uniform interpolation of the five features we use. This baseline performs worse than the strongest single feature filtering experiments by 0.5 to 1 BLEU points. For both the weight-based and feature-based methods of learning interpolation weights for the features, a significant number of \emph{candidate} runs are required before adequate performance is achieved. This is not surprising, since we are searching for an optimal weight vector in a fairly large weight space and we need a large number of samples before a good representation of the weight-reward function can be learned. Figure \ref{fig:candidate} shows the improvement in BLEU scores for the weight-based approach as data from more \emph{candidate} runs in added to the \emph{tuning} stage for learning weights and filtering the data set. The performance of the final NMT system steadily improves as more data from more systems is added and eventually converges.

Our strongest result was achieved with 14 \emph{candidate} runs for the weight-based approach for the 10, 15 and 20m setting respectively. This beat the uniform weight baseline by 1.5 to 2 BLEU points and the strongest single feature (LASER) baseline by 1 BLEU point. The feature based approach performed slightly better with 15 candidate runs and beat the strongest single feature baseline (LASER) by 1.3 BLEU points.

\section{Analysis} \label{sec:analysis}
The following sections examine the learned weights, the effect of transferring them to noisy corpora of a different language pair and the method's performance when exposed to specific kinds of noise. 

\subsection{Learned Weights}
Table \ref{tab:learned-weights} shows the weights learned using the \emph{tuning} network, normalized to sum to one. Unsurprisingly, the strongest feature (dual cross-entropy) has the highest weight, with the sentence length ratio and IBM Model 1 (weak multi-lingual features) drawn for the next place while source and target LM have relatively low weights. 
\begin{table}[ht]
    \centering
    \begin{tabular}{|c|c|c|} \hline
         \bf{Feature} & \bf{Weight} & \textbf{Feature} \\ \hline
         IBM Model 1 & 0.07 & 0.12 \\ \hline
         Source LM & 0.03 & 0.02 \\ \hline
         Target LM & 0.02 & 0.02 \\ \hline
         Dual xent & 0.81 & 0.76 \\ \hline
         Sen. Length Ratio & 0.07 & 0.08 \\ \hline
    \end{tabular}
    \caption{Feature weights learned post-tuning with the weight-based and the feature-based approaches. The weights have been normalized to sum to 1 (column).}
    \label{tab:learned-weights}
\end{table}

\subsection{Weight Transfer} \label{ssec:weight-transfer}
Since the feature functions we use for our experiments are reasonably language-independent, a reasonable experiment is to see if the feature weights learned on one language-pair can be transferred to a noisy corpus of another another language pair. However, we hypothesize that unless the feature distributions (proxy for noise profile of the dataset) of the datasets are similar, this transfer will have limited success.  

We test this hypothesis using the Maltese-English Paracrawl corpus. The training corpus contains 26.9 million sentence pairs and was sentence aligned using Vecalign and de-duplicated in a manner similar to our primary experiments. The validation and the test sets for these experiments are from the EUbookshop\footnote{\url{ opus.nlpl.eu/EUbookshop.php}} dataset and contain 3k and 2.2k sentences respectively. The sentence level features were computed using the procedure described in section ~\ref{ssec:features} and we use the DGT corpus\footnote{\url{data.europa.eu/euodp/en/data/dataset/dgt-translation-memory}} (about 1.6 million parallel sentences) to the train the clean translation models, the source and the target language models.

\begin{table}[ht!]
    \centering
    \begin{tabular}{|c|c|} \hline
         \multicolumn{2}{|c|}{\bf{1-Feature Filtering Baselines}} \\ \hline
        %  IBM Model 1 & - \\ \hline 
         Target LM & 28.3 \\ \hline
         Source LM & 27.1 \\ \hline
         Dual Cross-Entropy & \textbf{32.5} \\ \hline
        %  Sentence Length Ratio & - \\ \hline\hline
         \multicolumn{2}{|c|}{\bf{Filtering using Transfer Weights}} \\ \hline
         Uniform weight baseline & 30.5 \\ \hline
         Weight based & 31.6 \\ \hline
         Feature based & 31.3 \\ \hline
    \end{tabular}
    \caption{BLEU scores for the Maltese-English Paracrawl NMT systems where the training data was filtered using single features or a \emph{transferred} (from Estonian-English) weighted combination of features.}
    \label{tab:maltese-results}
\end{table}

The results of these experiments appear in Table~\ref{tab:maltese-results}. Even though filtering with the transferred weights beats the simpler single feature baselines, it fails to beat the strongest one, dual cross-entropy. It is worth noting that the reason filtering with the learned weights does this well is because the dual cross-entropy feature has the highest weight from our previous experiments. These experiments suggest that some form of feature distribution matching across corpora is required before weight transfer becomes viable.

\subsection{Sensitivity to Noise Types}
Inspired by \citet{khayrallah-koehn-2018-impact}, we look at how the most common noisy types in the Paracrawl data set affect the performance of the proposed method. For the purpose of these experiments, we use the \emph{Europarl} v8
\footnote{\url{www.statmt.org/europarl}}
Estonian-English data set. The training data set consists of about 651k parallel sentences, 11.2m source and 15.7m target tokens. We only use the feature-based method for this analysis and each experiment tunes weights based on 5 \emph{candidate} runs. 

We add synthetic noise to this data set by replacing 50\% of the sentences in the data set to contain a specific kind of noise. The noise types we looked at and their perturbation methods are described below:
\begin{enumerate}
    \item Misaligned sentences: Since parallel corpora extraction efforts use automated document and sentence alignment methods, noise includes source sentences which are not aligned to the correct target sentence. To emulate this, we randomly shuffle the source sentences of half the sentences in the clean data set. 
    \item Misordered words: A result of automatic or imperfect human translation, we add this noise to the clean data set by randomly shuffling the words within the source sentences.
    \item Wrong language: This is a very common noise type in web-crawled corpora. We emulate it by performing lexical replacements (from Estonian to French). 
    \item Untranslated words: This other common noise type is added to our data set by copying the source sentence to the target. 
\end{enumerate}

\begin{table}[t!]
    \centering
    \begin{tabular}{|c|c|} \hline
         \bf{Noise Type} & \bf{\% Retained} \\ \hline
         Misaligned sentences & 92 \\ \hline
         Misordered words & 81 \\ \hline
         Wrong language & 89 \\ \hline
         Untranslated words & 78 \\ \hline
    \end{tabular}
    \caption{The portion of the clean sentences retained after perturbing 50\% of the data set with specific noise types, learning feature weights and resampling the top 50\% samples.}
    \label{tab:noise-analysis}
\end{table}

For each type of noise, we perform the following experiment: perturb 50\% of the clean data with the chosen noise type, compute feature values for the sentences in the full data set, learn feature weights using the weight-based method described in section \ref{sec:methods}, filter out the top 50\% of the data set and measure the percentage of \emph{clean} (non-perturbed) sentences which were retained.\footnote{We note that the performance of this analysis depends on the chosen features. As an extreme example, if we perturb the source sentences and only consider a target-side feature (such as target language model scores), we will have no way of discriminating bad noisy samples from the clean ones.} The results of this analysis appears in Table \ref{tab:noise-analysis}. The method performs significantly better than chance in all noise categories, but given our choice of features, it is better at filtering out misaligned sentences and sentences with tokens in the wrong language and is slightly less effective at dealing with misordered and untranslated words.

\section{Future Work}
The validation set based delta-perplexity is expensive to compute per update and replacing it with a more stable or time invariant reward \cite{DBLP:journals/corr/abs-1911-10088} may help improve the performance of this method. Additionally, we plan to replace grid search with a more granular search procedure over the weight space with respect to the weight-feature-reward function. The tuning network can also be modified to include sentence-quality modulated loss functions (via feature values). An alternative to searching for feature weights is to instead search for the prototypical feature vector which maximizes translation performance and then use it to filter the closest sentence pairs from the noisy dataset. Finally, as discussed in Section~\ref{ssec:weight-transfer}, transferring learned weights has the potential to dramatically reduce the cost of applying to this method to new language pairs and may help with performance on low-resource language pairs where good feature weights cannot be learned.

\section{Conclusion}
We present a method for denoising and filtering noisy parallel data for improving the performance of neural machine translation systems. We learn interpolation weights for sentence-level features by modeling and searching over the weight-reward space. These are used to score and filter sentences in the noisy corpora. Our experiments with Estonian-English Paracrawl show gains of over a BLEU point over the strongest single feature filtering and uniform weight baselines. Analysis also shows that this method is effective at addressing the most common noise types in web-crawled corpora.

\clearpage

\bibliography{anthology,custom}

\begin{thebibliography}{29}
\expandafter\ifx\csname natexlab\endcsname\relax\def\natexlab#1{#1}\fi

\bibitem[{Antonova and Misyurev(2011)}]{antonova-misyurev-2011-building}
Alexandra Antonova and Alexey Misyurev. 2011.
\newblock \href {https://www.aclweb.org/anthology/W11-1218} {Building a
  web-based parallel corpus and filtering out machine-translated text}.
\newblock In \emph{Proceedings of the 4th Workshop on Building and Using
  Comparable Corpora: Comparable Corpora and the Web}, pages 136--144,
  Portland, Oregon. Association for Computational Linguistics.

\bibitem[{Ba{\~n}{\'o}n et~al.(2020)Ba{\~n}{\'o}n, Chen, Haddow, Heafield,
  Hoang, Espl{\`a}-Gomis, Forcada, Kamran, Kirefu, Koehn, Ortiz~Rojas,
  Pla~Sempere, Ram{\'\i}rez-S{\'a}nchez, Sarr{\'\i}as, Strelec, Thompson,
  Waites, Wiggins, and Zaragoza}]{banon-etal-2020-paracrawl}
Marta Ba{\~n}{\'o}n, Pinzhen Chen, Barry Haddow, Kenneth Heafield, Hieu Hoang,
  Miquel Espl{\`a}-Gomis, Mikel~L. Forcada, Amir Kamran, Faheem Kirefu, Philipp
  Koehn, Sergio Ortiz~Rojas, Leopoldo Pla~Sempere, Gema
  Ram{\'\i}rez-S{\'a}nchez, Elsa Sarr{\'\i}as, Marek Strelec, Brian Thompson,
  William Waites, Dion Wiggins, and Jaume Zaragoza. 2020.
\newblock \href {https://doi.org/10.18653/v1/2020.acl-main.417} {{P}ara{C}rawl:
  Web-scale acquisition of parallel corpora}.
\newblock In \emph{Proceedings of the 58th Annual Meeting of the Association
  for Computational Linguistics}, pages 4555--4567, Online. Association for
  Computational Linguistics.

\bibitem[{Bernier-Colborne and Lo(2019)}]{bernier-colborne-lo-2019-nrc}
Gabriel Bernier-Colborne and Chi-kiu Lo. 2019.
\newblock \href {https://doi.org/10.18653/v1/W19-5434} {{NRC} parallel corpus
  filtering system for {WMT} 2019}.
\newblock In \emph{Proceedings of the Fourth Conference on Machine Translation
  (Volume 3: Shared Task Papers, Day 2)}, pages 252--260, Florence, Italy.
  Association for Computational Linguistics.

\bibitem[{Brown et~al.(1993)Brown, Della~Pietra, Della~Pietra, and
  Mercer}]{brown-etal-1993-mathematics}
Peter~F. Brown, Stephen~A. Della~Pietra, Vincent~J. Della~Pietra, and Robert~L.
  Mercer. 1993.
\newblock \href {https://www.aclweb.org/anthology/J93-2003} {The mathematics of
  statistical machine translation: Parameter estimation}.
\newblock \emph{Computational Linguistics}, 19(2):263--311.

\bibitem[{Chaudhary et~al.(2019)Chaudhary, Tang, Guzm{\'a}n, Schwenk, and
  Koehn}]{chaudhary-etal-2019-low}
Vishrav Chaudhary, Yuqing Tang, Francisco Guzm{\'a}n, Holger Schwenk, and
  Philipp Koehn. 2019.
\newblock \href {https://doi.org/10.18653/v1/W19-5435} {Low-resource corpus
  filtering using multilingual sentence embeddings}.
\newblock In \emph{Proceedings of the Fourth Conference on Machine Translation
  (Volume 3: Shared Task Papers, Day 2)}, pages 261--266, Florence, Italy.
  Association for Computational Linguistics.

\bibitem[{ElNokrashy et~al.(2020)ElNokrashy, Hendy, Abdelghaffar, Afify,
  Tawfik, and Awadalla}]{elnokrashy2020score}
Muhammad~N. ElNokrashy, Amr Hendy, Mohamed Abdelghaffar, Mohamed Afify, Ahmed
  Tawfik, and Hany~Hassan Awadalla. 2020.
\newblock \href {http://arxiv.org/abs/2011.07933} {Score combination for
  improved parallel corpus filtering for low resource conditions}.

\bibitem[{Erdmann and Gwinnup(2019)}]{erdmann-gwinnup-2019-quality}
Grant Erdmann and Jeremy Gwinnup. 2019.
\newblock \href {https://doi.org/10.18653/v1/W19-5436} {Quality and coverage:
  The {AFRL} submission to the {WMT}19 parallel corpus filtering for
  low-resource conditions task}.
\newblock In \emph{Proceedings of the Fourth Conference on Machine Translation
  (Volume 3: Shared Task Papers, Day 2)}, pages 267--270, Florence, Italy.
  Association for Computational Linguistics.

\bibitem[{Esplà~Gomis et~al.(2020)Esplà~Gomis, SÃ¡nchez-Cartagena,
  Zaragoza-Bernabeu, and SÃ¡nchez-MartÃ­nez}]{esplgomis20a}
Miquel Esplà~Gomis, VÃ­ctor~M. SÃ¡nchez-Cartagena, Jaume
  Zaragoza-Bernabeu, and Felipe SÃ¡nchez-MartÃ­nez. 2020.
\newblock \href {https://www.aclweb.org/anthology/2020.wmt-1.107} {Bicleaner at
  wmt 2020: Universitat d'alacant-prompsit's submission to the parallel corpus
  filtering shared task}.
\newblock In \emph{Proceedings of the Fifth Conference on Machine Translation},
  pages 950--956, Online. Association for Computational Linguistics.

\bibitem[{Gonz{\'a}lez-Rubio(2019)}]{gonzalez-rubio-2019-webinterpret}
Jes{\'u}s Gonz{\'a}lez-Rubio. 2019.
\newblock \href {https://doi.org/10.18653/v1/W19-5437} {Webinterpret submission
  to the {WMT}2019 shared task on parallel corpus filtering}.
\newblock In \emph{Proceedings of the Fourth Conference on Machine Translation
  (Volume 3: Shared Task Papers, Day 2)}, pages 271--276, Florence, Italy.
  Association for Computational Linguistics.

\bibitem[{Hangya and Fraser(2018)}]{hangya-fraser-2018-unsupervised}
Viktor Hangya and Alexander Fraser. 2018.
\newblock \href {https://doi.org/10.18653/v1/W18-6477} {An unsupervised system
  for parallel corpus filtering}.
\newblock In \emph{Proceedings of the Third Conference on Machine Translation:
  Shared Task Papers}, pages 882--887, Belgium, Brussels. Association for
  Computational Linguistics.

\bibitem[{Junczys-Dowmunt(2018)}]{junczysdowmunt2019dual}
Marcin Junczys-Dowmunt. 2018.
\newblock \href {https://doi.org/10.18653/v1/W18-6478} {Dual conditional
  cross-entropy filtering of noisy parallel corpora}.
\newblock In \emph{Proceedings of the Third Conference on Machine Translation:
  Shared Task Papers}, pages 888--895, Belgium, Brussels. Association for
  Computational Linguistics.

\bibitem[{Khayrallah and Koehn(2018)}]{khayrallah-koehn-2018-impact}
Huda Khayrallah and Philipp Koehn. 2018.
\newblock \href {https://doi.org/10.18653/v1/W18-2709} {On the impact of
  various types of noise on neural machine translation}.
\newblock In \emph{Proceedings of the 2nd Workshop on Neural Machine
  Translation and Generation}, pages 74--83, Melbourne, Australia. Association
  for Computational Linguistics.

\bibitem[{Koehn et~al.(2007)Koehn, Hoang, Birch, Callison-Burch, Federico,
  Bertoldi, Cowan, Shen, Moran, Zens, Dyer, Bojar, Constantin, and
  Herbst}]{Koehn:2007:MOS:1557769.1557821}
Philipp Koehn, Hieu Hoang, Alexandra Birch, Chris Callison-Burch, Marcello
  Federico, Nicola Bertoldi, Brooke Cowan, Wade Shen, Christine Moran, Richard
  Zens, Chris Dyer, Ond\v{r}ej Bojar, Alexandra Constantin, and Evan Herbst.
  2007.
\newblock \href {http://dl.acm.org/citation.cfm?id=1557769.1557821} {Moses:
  Open source toolkit for statistical machine translation}.
\newblock In \emph{Proceedings of the 45th Annual Meeting of the ACL on
  Interactive Poster and Demonstration Sessions}, ACL '07, pages 177--180,
  Stroudsburg, PA, USA. Association for Computational Linguistics.

\bibitem[{Kumar et~al.(2019)Kumar, Foster, Cherry, and
  Krikun}]{kumar-etal-2019-reinforcement}
Gaurav Kumar, George Foster, Colin Cherry, and Maxim Krikun. 2019.
\newblock \href {https://doi.org/10.18653/v1/N19-1208} {Reinforcement learning
  based curriculum optimization for neural machine translation}.
\newblock In \emph{Proceedings of the 2019 Conference of the North {A}merican
  Chapter of the Association for Computational Linguistics: Human Language
  Technologies, Volume 1 (Long and Short Papers)}, pages 2054--2061,
  Minneapolis, Minnesota. Association for Computational Linguistics.

\bibitem[{Kurfal{\i} and {\"O}stling(2019)}]{kurfali-ostling-2019-noisy}
Murathan Kurfal{\i} and Robert {\"O}stling. 2019.
\newblock \href {https://doi.org/10.18653/v1/W19-5438} {Noisy parallel corpus
  filtering through projected word embeddings}.
\newblock In \emph{Proceedings of the Fourth Conference on Machine Translation
  (Volume 3: Shared Task Papers, Day 2)}, pages 277--281, Florence, Italy.
  Association for Computational Linguistics.

\bibitem[{Ng et~al.(2019)Ng, Yee, Baevski, Ott, Auli, and
  Edunov}]{ng-etal-2019-facebook}
Nathan Ng, Kyra Yee, Alexei Baevski, Myle Ott, Michael Auli, and Sergey Edunov.
  2019.
\newblock \href {https://doi.org/10.18653/v1/W19-5333} {{F}acebook {FAIR}{'}s
  {WMT}19 news translation task submission}.
\newblock In \emph{Proceedings of the Fourth Conference on Machine Translation
  (Volume 2: Shared Task Papers, Day 1)}, pages 314--319, Florence, Italy.
  Association for Computational Linguistics.

\bibitem[{Ott et~al.(2019)Ott, Edunov, Baevski, Fan, Gross, Ng, Grangier, and
  Auli}]{ott2019fairseq}
Myle Ott, Sergey Edunov, Alexei Baevski, Angela Fan, Sam Gross, Nathan Ng,
  David Grangier, and Michael Auli. 2019.
\newblock fairseq: A fast, extensible toolkit for sequence modeling.
\newblock In \emph{Proceedings of NAACL-HLT 2019: Demonstrations}.

\bibitem[{Parcheta et~al.(2019)Parcheta, Sanchis-Trilles, and
  Casacuberta}]{parcheta-etal-2019-filtering}
Zuzanna Parcheta, Germ{\'a}n Sanchis-Trilles, and Francisco Casacuberta. 2019.
\newblock \href {https://doi.org/10.18653/v1/W19-5439} {Filtering of noisy
  parallel corpora based on hypothesis generation}.
\newblock In \emph{Proceedings of the Fourth Conference on Machine Translation
  (Volume 3: Shared Task Papers, Day 2)}, pages 282--288, Florence, Italy.
  Association for Computational Linguistics.

\bibitem[{Rarrick et~al.(2011)Rarrick, Quirk, and Lewis}]{rarrick2011mt}
Spencer Rarrick, Chris Quirk, and Will Lewis. 2011.
\newblock \href
  {https://www.microsoft.com/en-us/research/publication/mt-detection-in-web-scraped-parallel-corpora/}
  {Mt detection in web-scraped parallel corpora}.
\newblock In \emph{Proceedings of MT Summit XIII}. Asia-Pacific Association for
  Machine Translation.

\bibitem[{Rossenbach et~al.(2018)Rossenbach, Rosendahl, Kim, Gra{\c{c}}a,
  Gokrani, and Ney}]{rossenbach-etal-2018-rwth}
Nick Rossenbach, Jan Rosendahl, Yunsu Kim, Miguel Gra{\c{c}}a, Aman Gokrani,
  and Hermann Ney. 2018.
\newblock \href {https://doi.org/10.18653/v1/W18-6487} {The {RWTH} {A}achen
  {U}niversity filtering system for the {WMT} 2018 parallel corpus filtering
  task}.
\newblock In \emph{Proceedings of the Third Conference on Machine Translation:
  Shared Task Papers}, pages 946--954, Belgium, Brussels. Association for
  Computational Linguistics.

\bibitem[{S\'{a}nchez-Cartagena et~al.(2018)S\'{a}nchez-Cartagena,
  Ba{\~n}\'{o}n, Ortiz-Rojas, and Ram\'{i}rez-S\'{a}nchez}]{prompsit:2018:WMT}
V\'{i}ctor~M. S\'{a}nchez-Cartagena, Marta Ba{\~n}\'{o}n, Sergio Ortiz-Rojas,
  and Gema Ram\'{i}rez-S\'{a}nchez. 2018.
\newblock Prompsit's submission to wmt 2018 parallel corpus filtering shared
  task.
\newblock In \emph{Proceedings of the Third Conference on Machine Translation,
  Volume 2: Shared Task Papers}, Brussels, Belgium. Association for
  Computational Linguistics.

\bibitem[{Schwenk and Douze(2017)}]{schwenk-douze-2017-learning}
Holger Schwenk and Matthijs Douze. 2017.
\newblock \href {https://doi.org/10.18653/v1/W17-2619} {Learning joint
  multilingual sentence representations with neural machine translation}.
\newblock In \emph{Proceedings of the 2nd Workshop on Representation Learning
  for {NLP}}, pages 157--167, Vancouver, Canada. Association for Computational
  Linguistics.

\bibitem[{Sen et~al.(2019)Sen, Ekbal, and
  Bhattacharyya}]{sen-etal-2019-parallel}
Sukanta Sen, Asif Ekbal, and Pushpak Bhattacharyya. 2019.
\newblock \href {https://doi.org/10.18653/v1/W19-5440} {Parallel corpus
  filtering based on fuzzy string matching}.
\newblock In \emph{Proceedings of the Fourth Conference on Machine Translation
  (Volume 3: Shared Task Papers, Day 2)}, pages 289--293, Florence, Italy.
  Association for Computational Linguistics.

\bibitem[{Soares and Costa-jussà(2019)}]{soares-2019-unsupervised}
Felipe Soares and Marta~R. Costa-jussà. 2019.
\newblock Unsupervised corpus filtering and mining.
\newblock In \emph{Proceedings of the Fourth Conference on Machine
  Translation}, Florence, Italy. Association for Computational Linguistics.

\bibitem[{Thompson and Koehn(2019)}]{thompson-koehn-2019-vecalign}
Brian Thompson and Philipp Koehn. 2019.
\newblock \href {https://doi.org/10.18653/v1/D19-1136} {{V}ecalign: Improved
  sentence alignment in linear time and space}.
\newblock In \emph{Proceedings of the 2019 Conference on Empirical Methods in
  Natural Language Processing and the 9th International Joint Conference on
  Natural Language Processing (EMNLP-IJCNLP)}, pages 1342--1348, Hong Kong,
  China. Association for Computational Linguistics.

\bibitem[{Venugopal et~al.(2011)Venugopal, Uszkoreit, Talbot, Och, and
  Ganitkevitch}]{venugopal-etal-2011-watermarking}
Ashish Venugopal, Jakob Uszkoreit, David Talbot, Franz Och, and Juri
  Ganitkevitch. 2011.
\newblock \href {https://www.aclweb.org/anthology/D11-1126} {Watermarking the
  outputs of structured prediction with an application in statistical machine
  translation.}
\newblock In \emph{Proceedings of the 2011 Conference on Empirical Methods in
  Natural Language Processing}, pages 1363--1372, Edinburgh, Scotland, UK.
  Association for Computational Linguistics.

\bibitem[{Wang et~al.(2018)Wang, Watanabe, Hughes, Nakagawa, and
  Chelba}]{wang-etal-2018-denoising}
Wei Wang, Taro Watanabe, Macduff Hughes, Tetsuji Nakagawa, and Ciprian Chelba.
  2018.
\newblock \href {https://doi.org/10.18653/v1/W18-6314} {Denoising neural
  machine translation training with trusted data and online data selection}.
\newblock In \emph{Proceedings of the Third Conference on Machine Translation:
  Research Papers}, pages 133--143, Brussels, Belgium. Association for
  Computational Linguistics.

\bibitem[{Wang et~al.(2019)Wang, Pham, Michel, Anastasopoulos, Neubig, and
  Carbonell}]{DBLP:journals/corr/abs-1911-10088}
Xinyi Wang, Hieu Pham, Paul Michel, Antonios Anastasopoulos, Graham Neubig, and
  Jaime~G. Carbonell. 2019.
\newblock \href {http://arxiv.org/abs/1911.10088} {Optimizing data usage via
  differentiable rewards}.
\newblock \emph{CoRR}, abs/1911.10088.

\bibitem[{Xu and Koehn(2017)}]{xu-koehn-2017-zipporah}
Hainan Xu and Philipp Koehn. 2017.
\newblock \href {https://doi.org/10.18653/v1/D17-1319} {{Z}ipporah: a fast and
  scalable data cleaning system for noisy web-crawled parallel corpora}.
\newblock In \emph{Proceedings of the 2017 Conference on Empirical Methods in
  Natural Language Processing}, pages 2945--2950, Copenhagen, Denmark.
  Association for Computational Linguistics.

\end{thebibliography}
\bibliographystyle{acl_natbib}

% \appendix

% \section{Weight transfer to other corpora}
% \label{sec:appendix}

\end{document}